\documentclass[10pt,journal]{IEEEtran}
\usepackage{amsmath,amsfonts}
\usepackage{algorithmic}
\usepackage{algorithm}
\usepackage{array}
\usepackage[caption=false,font=normalsize,labelfont=sf,textfont=sf]{subfig}
\usepackage{textcomp}
\usepackage{stfloats}
\usepackage{url}
\usepackage{verbatim}
\usepackage{graphicx}
\usepackage{orcidlink}
\usepackage{cite}
\usepackage{float}
\usepackage{booktabs}
\usepackage{arydshln}
\usepackage{ulem}
\usepackage{hyperref}
\usepackage{multirow}
\usepackage{graphicx}
\hypersetup{hidelinks=true}
\begin{document}

\title{UNetMamba: An Efficient UNet-Like Mamba for Semantic Segmentation of High-Resolution Remote Sensing Images}

\author{Enze~Zhu$^{\orcidlink{0009-0000-1717-5286}}$,~Zhan~Chen$^{\orcidlink{0009-0002-6039-0584}}$,~Dingkai~Wang$^{\orcidlink{0009-0009-3978-8054}}$,~Hanru~Shi,~Xiaoxuan~Liu$^{\orcidlink{0000-0002-1426-5056}}$,~and~Lei~Wang
\thanks{This work was supported in part by the Key Laboratory of Target Cognition and Application Technology under Grant 2023-CXPT-LC-005, and in part by the Science and Disruptive Technology Program under Grant AIRCAS2024-AIRCAS-SDTP-03. \textit{(Corresponding author: Xiaoxuan Liu.)}}
\thanks{E. Zhu, Z. Chen, D. Wang and H. Shi are with the Aerospace Information Research Institute, Chinese Academy of Sciences (CAS), Beijing 100094, China, Key Laboratory of Target Cognition and Application Technology, CAS, Beijing 100190, China, and the School of Electronic, Electrical and Communication Engineering, University of Chinese Academy of Sciences, Beijing 100049, China (e-mail: \{zhuenze23, chenzhan21, wangdingkai23, shihanru23\}@mails.ucas.ac.cn).

X. Liu and L. Wang are with the Aerospace Information Research Institute, CAS, Beijing 100094, China, and Key Laboratory of Target Cognition and Application Technology, CAS, Beijing 100190, China (e-mail: \{liuxiaoxuan, wanglei002931\}@aircas.ac.cn).}}

\markboth{Journal of \LaTeX\ Class Files,~Vol.~xx, No.~xx, August~2024}%
{Shell \MakeLowercase{\textit{et al.}}: A Sample Article Using IEEEtran.cls for IEEE Journals}

\IEEEpubid{}

\maketitle

\begin{abstract}
Semantic segmentation of high-resolution remote sensing images is vital in downstream applications such as land-cover mapping, urban planning and disaster assessment. Existing Transformer-based methods suffer from the constraint between accuracy and efficiency, while the recently proposed Mamba is renowned for being efficient. Therefore, to overcome the dilemma, we propose UNetMamba, a UNet-like semantic segmentation model based on Mamba. It incorporates a mamba segmentation decoder (MSD) that can efficiently decode the complex information within high-resolution images, and a local supervision module (LSM), which is train-only but can significantly enhance the perception of local contents. Extensive experiments demonstrate that UNetMamba outperforms the state-of-the-art methods with mIoU increased by 0.87\% on LoveDA and 0.39\% on ISPRS Vaihingen, while achieving high efficiency through the lightweight design, less memory footprint and reduced computational cost. The source code is available at \href{https://github.com/EnzeZhu2001/UNetMamba}{https://github.com/EnzeZhu2001/UNetMamba}.
\end{abstract}


\section{Introduction}
\IEEEPARstart{S}purred by the progress in advanced imaging techniques, high-resolution remote sensing images have become increasingly accessible, which can provide richer contents than regular images. Such intricate contents render them crucial for semantic segmentation in several downstream applications, yet simultaneously present significant challenges regarding accuracy and efficiency due to the high complexity of information.

As the foundation models of computer vision continuously evolve, research on semantic segmentation of high-resolution remote sensing images has flourished. The CNN-based method UNet\cite{ref1} is a milestone in semantic segmentation, which has established the foundational architecture for subsequent works: an encoder-decoder framework with skip connections. With the success of Transformer\cite{ref2}, Wang et al.\cite{ref3} introduce it into the U-shape framework, significantly improving the semantic segmentation accuracy of high-resolution remote sensing images. Consequently, numerous subsequent works have focused on innovating Transformer-based methods\cite{ref4, ref5, ref6}. Despite the prosperity of Transformer in remote sensing semantic segmentation, its quadratic computational complexity and high parameter count\cite{ref7} have inevitably limited the application efficiency in high-resolution images.

Recently, Mamba \cite{ref8} has been proposed and attracted extensive attention. For efficiency, it achieves a linear complexity, while guaranteeing accuracy through its competitive long-distance dependency modeling capability. Subsequently, Mamba-based visual foundation models\cite{ref9, ref10} have been well-designed. Benefiting from such models, a series of Mamba-based UNets\cite{ref11, ref12, ref13} have achieved impressive accuracy in medical imagery segmentation, showing the potential of visual Mamba in segmentation. For remote sensing imagery interpretation, visual Mamba has also demonstrated promising efficiency, with RSMamba\cite{ref14} for classification, CDMamba\cite{ref15} for change detection, and RS$^3$Mamba\cite{ref18} for semantic segmentation. However, these pioneering works have primarily validated the effectiveness of Mamba in their respective tasks, without fully leveraging its dual advantages in accuracy and efficiency through targeted design.

To overcome the dilemma between accuracy and efficiency and then achieve efficient semantic segmentation of high-resolution remote sensing images, we propose a Mamba-based UNet-like model named UNetMamba. It consists of three main components: an encoder using ResT\cite{ref19} backbone, a mamba segmentation decoder (MSD) for efficiently decoding, and a neatly-designed local supervision module (LSM) to enhance the perception of local semantic information. The contributions of our research are summarized as follows:

1) We devise MSD, a plug-and-play Mamba-based decoder for efficiently decoding semantic information from multi-scale feature maps. Specifically, visual state space (VSS) blocks of VMamba\cite{ref9} are expressly transferred to the decoding side,  significantly reducing the parameter count while leveraging the long-distance modeling capability to accurately decode complex information in a global receptive field.

2) We also conceive a CNN-based LSM to address the lack of local information perception in MSD. With two different scales of convolutions and an auxiliary loss function, we enhance the perception of local semantic information through supervised training. The train-only design also saves on inference costs, further ensuring the application efficiency.

3) Based on MSD and LSM, we propose UNetMamba, a UNet-like model for efficient semantic segmentation of high-resolution remote sensing images. It has the advantages of being lightweight and requiring low costs, while also achieving state-of-the-art (SOTA) performance on two well-known high-resolution remote sensing imagery datasets.

\section{Methodology}
\begin{figure*}
    \centering
    \includegraphics[width=1\linewidth]{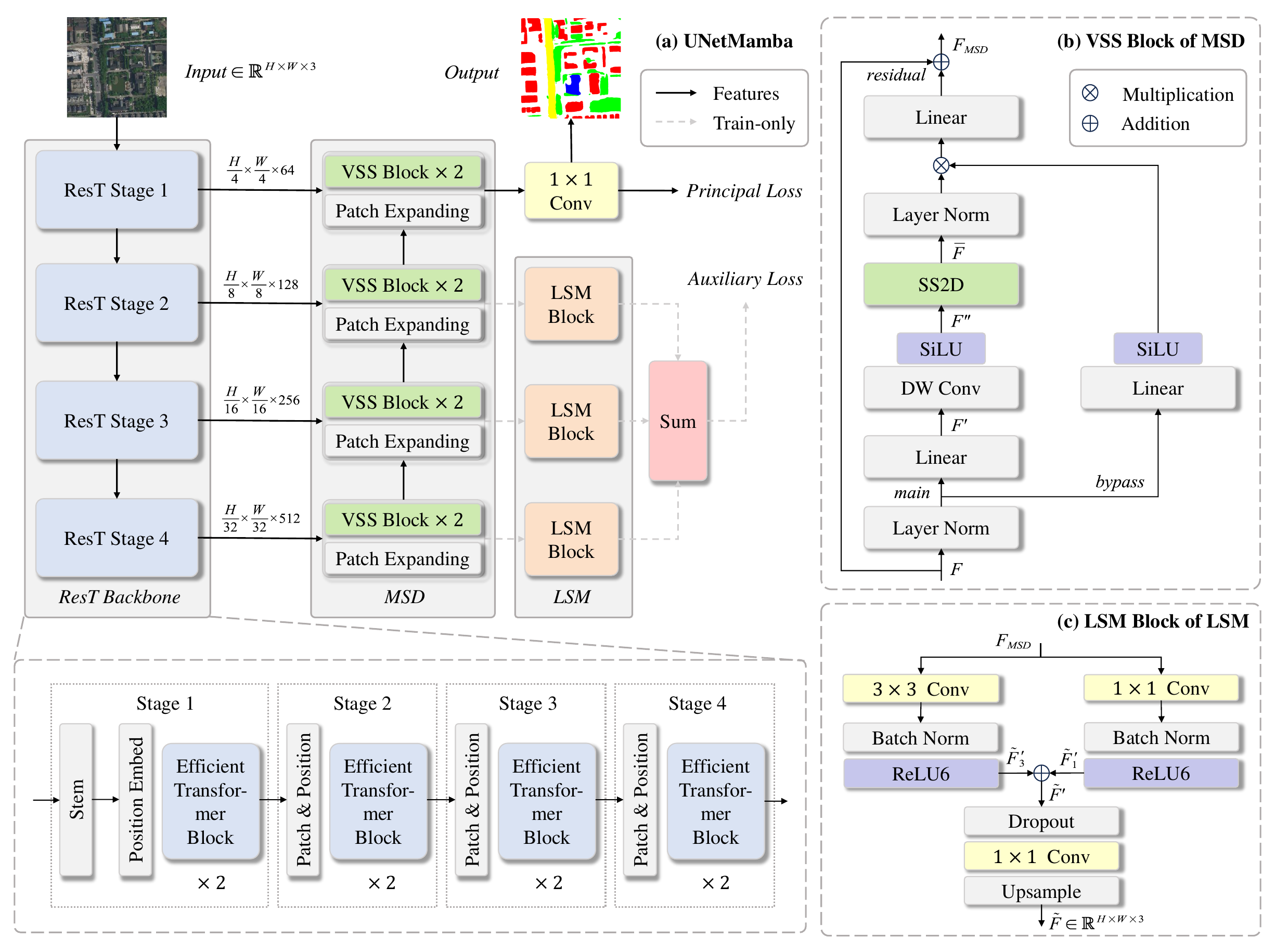}
    \caption{Framework of the proposed UNetMamba. (a) Overall architecture of UNetMamba. (b) Visual State Space (VSS) block of Mamba Segmentation Decoder (MSD). (c) Local Supervision Module (LSM) Block of LSM. }
    \label{fig:1}
\end{figure*}

As shown in Fig. \ref{fig:1}(a), the overall architecture of UNetMamba is constructed under a U-shape framework \cite{ref1}, where a pre-trained ResT backbone\cite{ref19} is selected as encoder while decoder is the proposed MSD. Moreover, the LSM is designed to enhance the perception of local semantic information.

\subsection{ResT Encoder}
As illustrated in Fig. \ref{fig:1}(a), with Efficient Transformer Block (ETB) as the core, ResT encoder is consisted of four stages to capture multi-scale feature maps. To avoid the quadratic computational costs of vanilla multi-head self-attention, ETB employs efficient multi-head self-attention (EMSA)\cite{ref19}:
\begin{equation}
\label{emsa}
\resizebox{.9\hsize}{!}{$\text{EMSA}(Q, K, V) = \text{IN}\left(\text{Softmax}\left(\text{Conv}\left(\frac{Q K^T}{\sqrt{d_k}}\right)\right)\right)V$}
\end{equation}
where an instance
normalization IN(·) and a 1×1 convolution Conv(·) are introduced to improve efficiency. When applied to high-resolution remote sensing images, EMSA can ensure the high encoding efficiency of our UNetMamba.

\subsection{Mamba Segmentation Decoder}
Since high-resolution remote sensing images can be treated as very long sequences when flattened, the linear scaling ability of Mamba\cite{ref8}  has a natural advantage in processing them compared with the quadratic one of Transformer\cite{ref2}.

Actually, in pioneering studies\cite{ref11, ref12, ref13, ref18}, the benefits of employing Mamba for semantic segmentation have been established. However, models that utilize a pre-trained VMamba\cite{ref9}  backbone ($>$30M) while encoding tend to be overweight for our goal of high efficiency. Therefore, we incorporate the basic unit of VMamba, namely, VSS block\cite{ref9}, into the decoding side only.

The structure of a VSS block is depicted in Fig. \ref{fig:1}(b). The 2-D feature map $F$, which has undergone patch expansion, firstly goes through layer normalization and then splits into two flows. The main flow is further linearly embedded as $F'$:
\begin{equation}
\label{e-1}
F' = \text{LinearEmbed}\left( \text{LayerNorm}\left( \text{PatchExp}\left( F \right) \right) \right)
\end{equation}
Subsequently, $F'$ undergoes a 3×3 depth-wise convolution and is then SiLU-activated into $F''$:
\begin{equation}
\label{e-2}
F'' = \text{SiLU}\left( \text{DWConv}\left( F' \right) \right)
\end{equation}
The 2-D Selective Scan (SS2D) module thereafter scans $F''$ in four different directions, decoding semantic information under a global receptive field and linear complexity:
\begin{equation}
\label{e-3}
\begin{split}
F_v &= \text{ScanExp}\left( F'', v \right), \quad v \in \left\{ 1,2,3,4 \right\} \\
\bar{F}_v &= \text{S6}\left( F_v \right), \quad v \in \left\{ 1,2,3,4 \right\} \\
\bar{F} &= \text{ScanMerge}\left( \bar{F}_1, \bar{F}_2, \bar{F}_3, \bar{F}_4 \right)
\end{split}
\end{equation}
where the expansion and merging operation correspond to VMamba\cite{ref9}, while the S6 operation denotes the selective scan space state sequential model of Mamba\cite{ref8}. The $\bar{F}$ then goes through layer normalization and an element-wise multiplication with the bypass flow, which has also undergone the same linear embedding and Silu activation. Finally, the VSS block outputs $F_{MSD}$ through a linear layer with residual connection.

Based on aforementioned operations, the MSD decodes multi-scale feature maps at four different stages, ultimately outputting the semantic segmentation results through a 1×1 convolution head.

\begin{table*}[!htbp]
\caption{\textbf{Quantitative Comparison Results on the LoveDA Dataset at a size of 1024 × 1024 pixels. The best values are in bold.}}
\label{tb1}
\centering
\begin{tabular}{cccccccccc}
\toprule
\textbf{Method}& \textbf{Backbone} &\textbf{Background}&\textbf{Building}&\textbf{Road}&\textbf{Water}&\textbf{Barren}&\textbf{Forest}&\textbf{Agriculture}&\textbf{mIoU(\%)} \\
\midrule
         BANet\cite{ref3}&       ResT-Lite&41.92&  54.13&  52.39&  76.18&  11.74&  45.07&  51.71&  47.59 \\
         MANet\cite{ref5}&     ResNet-50&44.38&  56.20&  56.03&  78.41&  17.06&  45.81&  58.17&  50.87 \\
         DC-Swin\cite{ref6}&     Swin-T&40.75&  57.81&  56.55&  79.59&  16.58&  46.49&  57.76&  50.79 \\
 UNetFormer\cite{ref7}& ResNet-18&45.47& 58.75& 56.53& 80.33& \textbf{18.90}& 45.46& 61.92& 52.48\\
 E-PyramidMamba\cite{ref16}& ResNet-18& 45.60& 56.37& 54.26& 80.33& 15.77& 46.06& 61.73&51.45\\
 CM-UNet\cite{ref17}& ResNet-18& 43.57& 55.59& 52.85& 77.80& 16.19& 43.15& 58.36&49.64\\
         RS$^3$Mamba\cite{ref18}&   R18-VMamba-T&41.60&  58.23&  54.03&  77.34&  17.97&  43.81&  61.37&  50.62\\
\midrule
         UNetMamba(ours)&  ResT-Lite&\textbf{47.08}&  \textbf{59.16}&  \textbf{56.74}&  \textbf{81.37}&  18.15&  \textbf{46.61}&  \textbf{64.31}&  \textbf{53.35}\\
         
\bottomrule
\end{tabular}
\end{table*}

\begin{table*}[!htbp]
\caption{\textbf{Quantitative Comparison Results on the ISPRS Vaihingen Dataset at a size of 1024 × 1024 pixels.\\
The metrics are measured on a single NVIDIA RTX 4090. The best values are in bold.}}
\label{tb2}
\centering
\renewcommand{\arraystretch}{1}
\begin{tabular}{cccccccc}
\toprule
\textbf{Method}&\textbf{Backbone}&\textbf{Param(M)}&\textbf{Memo(MB)}&\textbf{FLOPs(G)}&\textbf{mF1(\%)}&\textbf{mIoU(\%)}&\textbf{OA(\%)}\\
\midrule
BANet\cite{ref3}& ResT-Lite&12.73&194.61  &85.43& 90.32& 82.45&91.92\\
MANet\cite{ref4}& ResNet-50& 35.86 &547.87  &216.82& 90.68& 83.06&92.28\\
DC-Swin\cite{ref5}&Swin-T&45.63 &694.62  &190.04&  90.71&  83.08& 92.30\\
 UNetFormer\cite{ref7}& ResNet-18& \textbf{11.72}& \textbf{179.34}  &\textbf{47.03}& 90.59& 82.93&92.21\\
 E-PyramidMamba\cite{ref16}& ResNet-18& 28.76& 439.15& 75.97& 90.74& 83.08&92.29\\
 CM-UNet\cite{ref17}& ResNet-18& 13.11& 199.96& 48.66& 90.32& 82.60&92.02\\  
RS$^3$Mamba \cite{ref18}&R18-VMamba-T&43.32 &662.21  &157.89&  90.73&  83.08& 92.27\\
\midrule
UNetMamba(ours)&ResT-Lite&14.76&225.71&100.52&  \textbf{90.95}&  \textbf{83.47}& \textbf{92.51}\\
\bottomrule
\end{tabular}
\end{table*}

\subsection{Local Supervision Module}
The large receptive field of SS2D in VSS blocks tend to be double-edged, causing MSD to partially overlook local semantic information while decoding. However, such details in high-resolution images are key factors to further improve the accuracy of semantic segmentation. Therefore, a CNN-based flexible LSM is proposed to enhance the perception of local details, as depicted in Fig. \ref{fig:1}(c). 

Considering the relatively smaller scale of land-covers in high-resolution scenarios, we adopt two parallel convolution branches in LSM with the kernel size set as 3 and 1, respectively. Each branch is followed by a batch normalization layer with a ReLU6 activation:
\begin{equation}
\label{e-4}
\tilde{F'}_i=\text{ReLU6} \left( \text{BatchNorm}\left(\text{Conv}_i\left( F_{MSD} \right) \right) \right)
\end{equation}
where $i \in \left\{1,3\right\}$ denotes the kernel size, while $F_{MSD}$ denotes the output feature of MSD at the corresponding stage. The two branches then merge into $\tilde{F'}=\tilde{F}_{1}^{'}\oplus \tilde{F}_{3}^{'}$.

Eventually, through a dropout layer, a 1×1 convolution consistent with the segmentation head in MSD, and an upsampling layer to the original image size, LSM obtains $\tilde{F}\in \mathbb{R}^{H\times W\times 3}$:
\begin{equation}
\label{e-6}
\tilde{F}=\text{Upsample}\left(\text{Conv}_1\left(\text{Dropout}\left( \tilde{F'}\right) \right) \right)
\end{equation}

During training, we add LSM blocks with different input channels to the decoder at stage 2-4 and then attain the segmentation result by summing up, which is further used in calculating the auxiliary loss function.

\subsection{Loss Function}
The loss function utilized in training UNetMamba consists of two parts, a principal loss $L_{p}$ for overall optimization and an auxiliary loss $L_{a}$ for local supervision.

To guarantee the holistic segmentation performance of UNetMamba, two classic loss functions are selected to jointly form the principal loss $L_{p}$:
\begin{equation}
\label{e-7}
L_{dice}=1-\frac{2}{N}\sum_{n=1}^N{\sum_{k=1}^K{\frac{\hat{y}_{k}^{\left( n \right)}y_{k}^{\left( n \right)}}{\hat{y}_{k}^{\left( n \right)}+y_{k}^{\left( n \right)}}}}
\end{equation}
\begin{equation}
\label{e-8}
L_{ce}=-\frac{1}{N}\sum_{n=1}^N{\sum_{k=1}^K{\hat{y}_{k}^{\left( n \right)}\log y_{k}^{\left( n \right)}}}
\end{equation}
where $L_{dice}$ and $L_{ce}$ represents dice loss and cross-entropy loss, respectively. And $N$ denotes the number of samples while $K$ denotes the number of categories. $\hat{y}_{k}^{\left( n \right)}$ is the $k$-th element in one-hot encoding of the true label for sample $n$, while $y_{k}^{\left( n \right)}$ is the confidence of sample $n$ belonging to category $k$. 

Moreover, in order to efficiently achieve the local supervisory role of LSM, the most classic loss function in semantic segmentation $L_{ce}$ \cite{ref1} is also used as the auxiliary loss $L_{a}$. 

Eventually, to balance the effects of $L_{p}$ and $L_{a}$ and then achieve optimal performance, the overall loss function for training UNetMamba is formulated as a weighted equation: 
\begin{equation}
\label{e-9}
L=L_{p}+\alpha L_{a}=(L_{dice}+L_{ce})+\alpha L_{ce}
\end{equation}
where the weight factor $\alpha$ is set as 0.4\cite{ref6}.

\begin{figure}
    \centering
    \includegraphics[width=1\linewidth]{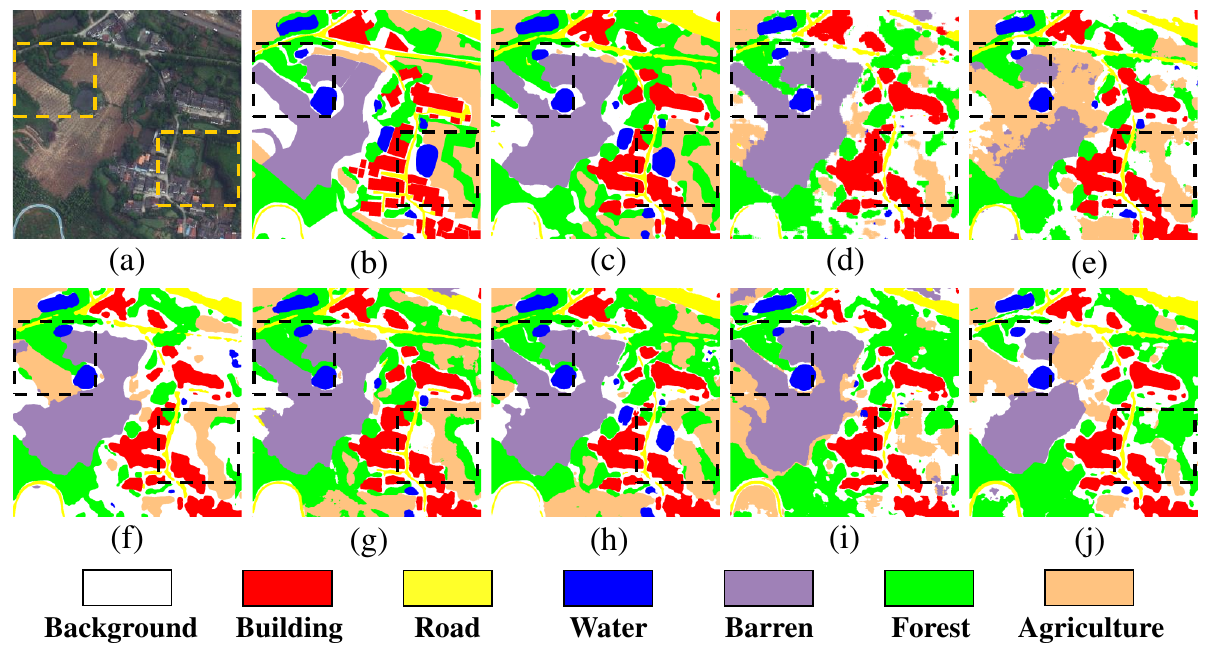}
    \caption{Qualitative comparison on the LoveDA dataset at resolution of 1024 × 1024 pixels. (a) Origin, (b) Ground Truth, (c) the proposed UNetMamba, (d) BANet, (e) MANet, (f) DC-Swin, (g) UNetFormer, (h) E-PyramidMamba, (i) CM-UNet and (j) RS$^3$Mamba.}
    \label{fig:2}
\end{figure}
\begin{figure}
    \centering
    \includegraphics[width=1\linewidth]{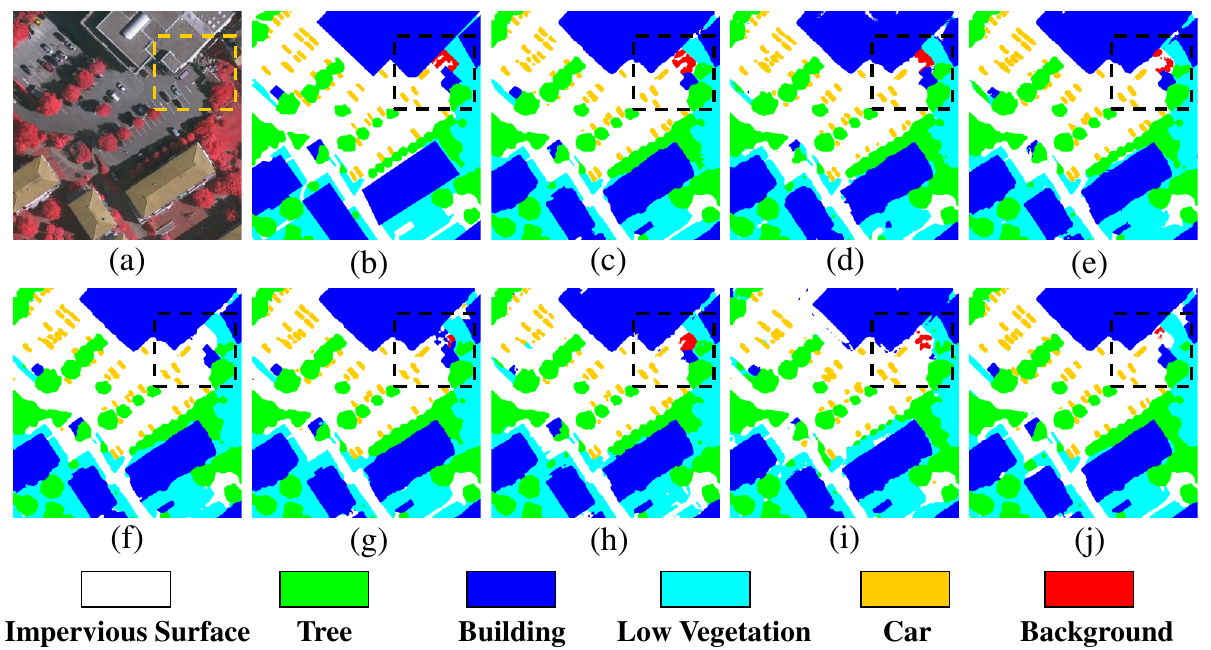}
    \caption{Qualitative comparison on the ISPRS Vaihingen dataset at resolution of 1024 × 1024 pixels. (a) Origin, (b) Ground Truth, (c) the proposed UNetMamba, (d) BANet, (e) MANet, (f) DC-Swin, (g) UNetFormer, (h) E-PyramidMamba, (i) CM-UNet and (j) RS$^3$Mamba.}
    \label{fig:3}
\end{figure}

\section{Experimental results}
\subsection{Datasets}
\textit{1) LoveDA}: The LoveDA dataset\cite{ref20} contains 5987 high-resolution remote sensing images at the same size of 1024×1024 pixels and includes 7 land-cover categories (background, building, road, water, barren, forest and agriculture). The dataset encompasses two scenes (urban and rural) collected from three Chinese cities, which brings considerable challenges due to multi-scale land-covers and complex contents. In our experiments, following the official settings\cite{ref20}, 2522 images were utilized for training, 1669 images for validation and 1796 images for testing without further cropping. 

\textit{2) ISPRS Vaihingen}: The ISPRS Vaihingen dataset contains 33 TOP (True OrthoPhoto) images, each with very high spatial resolution at an average size of 2494×2064 pixels. This widely-used remote sensing dataset involves 6 land-cover categories (impervious surface, tree, building, low vegetation, car and background). In the unbiased experiments, we randomly selected 9 images for testing, 2 images for validating with the remaining images for training, which was repeated for 5 times, and cropped all the 5 different sets of images into patches at a size of 1024×1024 pixels, respectively.

\subsection{Experimental Settings}
All of the experiments were implemented with PyTorch on a single NVIDIA RTX 4090 GPU, and the optimizer was set as AdamW with a 0.0006 learning rate and a 0.00025 weight decay. For both of the datasets, the training epoch was set as 100 while the batch size was 8. 

Moreover, the number of model parameters (Param), memory footprint (Memo) and floating point operation count (FLOPs) were selected for efficiency evaluation, while the mean F1-score (mF1), mean intersection over union (mIoU) and overall accuracy (OA) were chosen as accuracy evaluation metrics.

\subsection{Performance Comparison}
We conducted a comprehensive performance comparison on LoveDA and ISPRS Vaihingen datasets. A series of SOTA optical remote sensing imagery semantic segmentation models were selected as our competitors, including Transformer-based models: BANet\cite{ref3}, MANet\cite{ref4}, DC-Swin\cite{ref5}, UNetFormer\cite{ref6}, and Mamba-based models: Efficient PyramidMamba (E-PyramidMamba)\cite{ref16}, CM-UNet\cite{ref17}, RS$^3$Mamba\cite{ref18}.
 
\textit{1) Comparison on the LoveDA Dataset}: As shown in Tab. \ref{tb1}, the proposed UNetMamba achieved the best accuracy performance in quantitative comparison, with a significant improvement of 0.87\% in mIoU. It is note-worthy that our UNetMamba achieved SOTA in six out of seven categories. In background and agriculture, UNetMamba led the second place by 1.48\% and 2.39\%, respectively. This distinct advantage indicated that, benefiting from MSD, our UNetMamba garnered excellent global perception ability, which made it capable of accurately segmenting large land-covers under such a high-resolution condition. The outstanding performance of UNetMamba in building, road and other small land-cover categories proved that the train-only LSM successfully enhanced the perception of  local contents through supervision and generalization. The qualitative comparison results on LoveDA are illustrated in Fig. \ref{fig:2}, which intuitively shows that the semantic segmentation mask obtained by UNetMamba is more accurate with less omission and clearer boundaries.

\textit{2) Comparison on the ISPRS Vaihingen Dataset}: The proposed UNetMamba not only achieved SOTA performance on the challenging LoveDA dataset, but also performed noticeably well on the classic Vaihingen dataset. As shown in Tab. \ref{tb2}, in terms of accuracy, our UNetMamba achieved improvements of 0.21\% in mF1, 0.39\% in mIoU, and 0.21\% in OA, while maintaining competitive efficiency through a lightweight model (14.76M) with low costs (225.71MB, 100.52G). Fig. \ref{fig:3} illustrates the qualitative comparison results, with the framed areas further demonstrating the competitiveness of UNetMamba in the perception of local semantic details, which is particularly important for the semantic segmentation accuracy in high-resolution remote sensing scenes.

\begin{table}
\caption{\textbf{Ablation Study results on the LoveDA and ISPRS Vaihingen  Dataset. The metrics are measured by a 1024 × 1024 input on a single NVIDIA RTX 4090, and the best values are in bold.}}
\label{tb3}
\centering
\renewcommand\arraystretch{1.25}
\setlength{\tabcolsep}{1.2mm}{
    \begin{tabular}{cc|cccc}
    \hline
          \multirow{2}{*}{\textbf{\quad MSD}}&  \multirow{2}{*}{\textbf{LSM}}&\multirow{2}{*}{\textbf{Param(M)}}&\multirow{2}{*}{\textbf{FLOPs(G)}}& \multicolumn{2}{c}{\textbf{mIoU(\%)}} \\
          \cline{5-6}
          & & & & \textit{LoveDA} & \textit{Vaihingen}\\
    \hline
         \multicolumn{2}{c|}{\textit{Baseline with ResT-Lite}}& 42.14&183.64&52.64 & 83.23\\
    \hline
         \quad\checkmark&  &\textbf{13.89}&\textbf{100.52}&52.61 & 83.18\\
         \quad\checkmark& \checkmark &14.76&\textbf{100.52}&\textbf{53.35} & \textbf{83.47}\\
    \hline
    \end{tabular}}
\end{table}
\subsection{Ablation Studies}
To assess the effectiveness of the proposed MSD and LSM in UNetMamba, ablation studies were conducted on both of the datasets, with the results listed in Tab. \ref{tb3}. For the sake of credibility, the baseline model, RS$^3$Mamba\cite{ref17}, was updated by replacing its main encoding branch with the ResT\cite{ref18} backbone consistent with UNetMamba. And after removing its Mamba-based auxiliary branch and upgrading its decoder into our MSD, the model experienced significant reduction in Param and FLOPs, with only minor drop in mIoUs, demonstrating the high efficiency of MSD in decoding complex semantic information. On both datasets, deployment of LSM also led to 0.74\% and 0.29\% increase in mIoUs, respectively. However, the cost was merely a modest parameter-count increase of 0.87M and a constant computation complexity due to the train-only scheme, which further confirming the high efficiency of LSM in enhancing the perception of local semantic information.

\section{Conclusion}
In this paper, an efficient semantic segmentation model called UNetMamba is proposed for high-resolution remote sensing images. Considering the multi-scale land-covers and complex information in such images, our UNetMamba features a plug-and-play Mamba-based segmentation decoder MSD for efficient semantic information decoding, and a train-only local supervision module LSM that can efficiently enhance the perception of local semantic details. Extensive experiments conducted on two well-known remote sensing datasets demonstrate that UNetMamba not only outperforms other SOTA models, but also achieves light weight and low costs. In the future, we will continue to explore promising linear mechanisms to further improve the accuracy and efficiency of our UNetMamba.

\vfill

\end{document}